\documentclass[letterpaper, 10 pt, conference]{ieeeconf}  
\IEEEoverridecommandlockouts                              
\usepackage{graphics} 
\usepackage{epsfig} 
\usepackage{mathptmx} 
\usepackage{times} 
\usepackage{amsmath,amssymb,amsfonts}
\usepackage{siunitx} 
\usepackage{textcomp}
\usepackage{gensymb} 
\usepackage{booktabs}
\usepackage{multirow}
\usepackage{xcolor}

\let\labelindent\relax
\usepackage{enumitem}
\usepackage{xspace}

\usepackage{cite}
\usepackage{makecell}
\usepackage{multicol}
\usepackage{rotating}
\usepackage[percent]{overpic}
\usepackage{contour}
\usepackage{courier}

\usepackage{subcaption} 
\usepackage[font=small]{caption}
\usepackage[percent]{overpic}

\usepackage[11pt]{moresize}

\usepackage{pifont}
\definecolor{MyDarkBlue}{rgb}{0,0.08,1}
\definecolor{airforceblue}{rgb}{0.36, 0.54, 0.66}
\definecolor{MyDarkGreen}{rgb}{0.02,0.6,0.02}
\definecolor{MyDarkRed}{rgb}{0.8,0.02,0.02}
\definecolor{MyDarkOrange}{rgb}{0.40,0.2,0.02}
\definecolor{MyPurple}{RGB}{111,0,255}
\definecolor{MyRed}{rgb}{1.0,0.0,0.0}
\definecolor{MyGold}{rgb}{0.75,0.6,0.12}
\definecolor{MyDarkgray}{rgb}{0.66, 0.66, 0.66}
\definecolor{MyPink}{rgb}{0.9, 0.33, 0.5}
\definecolor{MyCyan}{rgb}{0., 0.4, 0.4}

\usepackage{tikz}

\usepackage{tikz}

\usepackage{algorithm}
\usepackage{algorithmic}
\usepackage{graphicx}
\usepackage[symbol]{footmisc}

\usepackage[T1]{fontenc}
\usepackage[utf8]{inputenc}

\usepackage[pagebackref=true,breaklinks=true,bookmarks=true,colorlinks]{hyperref}
\def\BibTeX{{\rm B\kern-.05em{\sc i\kern-.025em b}\kern-.08em
T\kern-.1667em\lower.7ex\hbox{E}\kern-.125emX}}

\title{\LARGE \bf
Out-of-Distribution Recovery with Object-Centric Keypoint Inverse \\  Policy for Visuomotor Imitation Learning
}

\author{George Jiayuan Gao$^{*}$, Tianyu Li, and Nadia Figueroa \\ 
University of Pennsylvania\\
}

\begin{document}

\twocolumn[{%
    \renewcommand\twocolumn[1][]{#1}%
    \maketitle
    \vspace{-5pt}
    \begin{center}
            \includegraphics[trim={0cm 8cm 0cm 8cm},clip,width=0.93\textwidth]{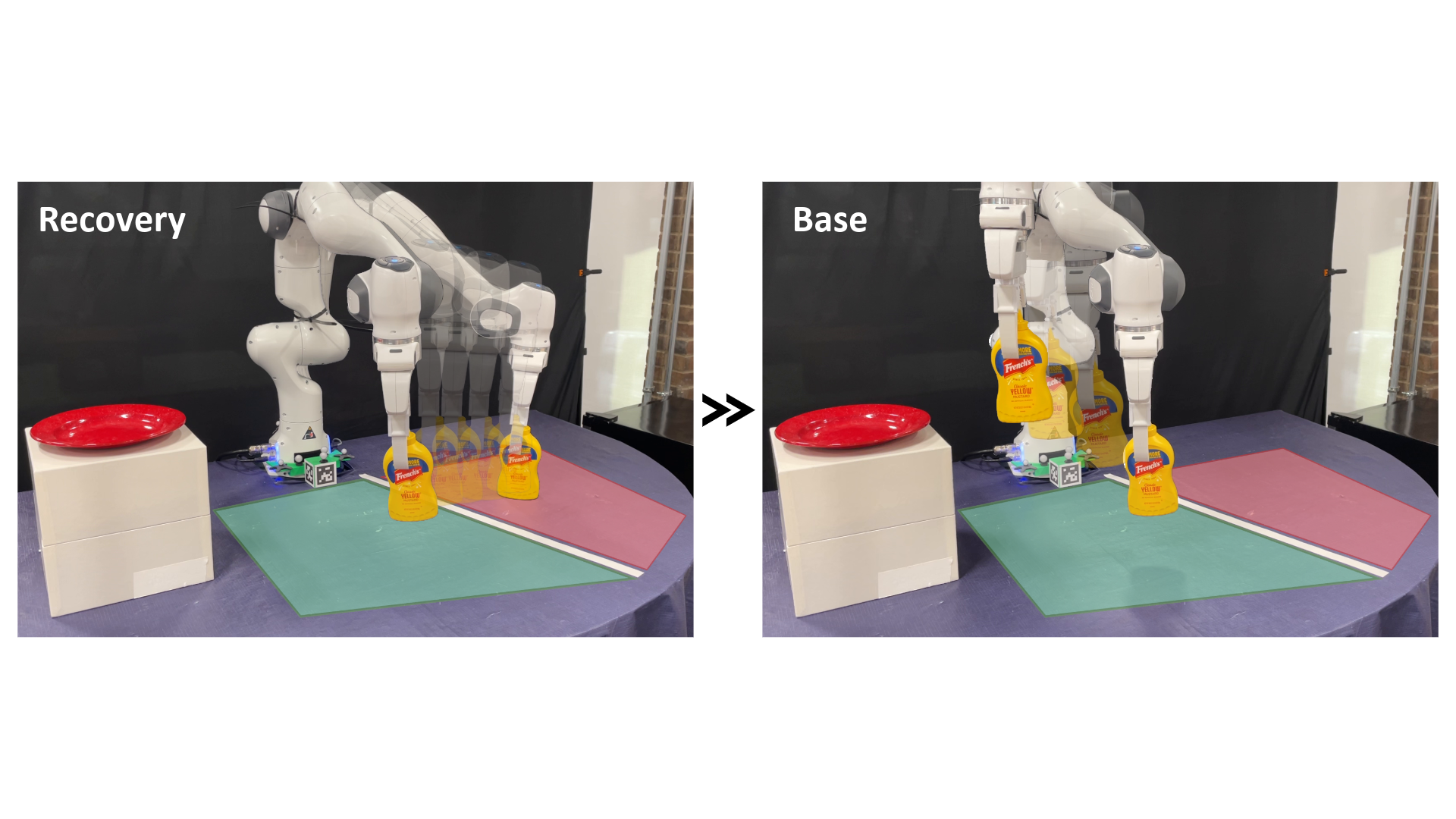} 
            \captionof{figure}{\textbf{Object-Centric Recovery (OCR) on Bottle Pick and Place Task.} \label{fig:leading} The base visuomotor policy (\textbf{Right}), trained on the bottle’s initial pose within the green-shaded region of the table, exhibits limited generalization when the bottle is initialized in the red-shaded region, which is considered out-of-distribution (OOD). (\textbf{Left}) showed the recovery policy using our OCR framework to recover from the red-shaded OOD region, returning the system to a region of high confidence for the base visuomotor policy, where it resumes control.} 
            \vspace{2mm}
    \end{center}}]

\footnote[0]{$^{*}$Corresponding author. (e-mail: gegao@seas.upenn.edu)}
\footnote[0]{All authors are with School of Engineering and Applied Science, University of Pennsylvania, Pennsylvania, PA 19104 USA.}

\vspace{-9pt}
\begin{abstract}
We propose an object-centric recovery (OCR) framework to address the challenges of out-of-distribution (OOD) scenarios in visuomotor policy learning. Previous behavior cloning (BC) methods rely heavily on a large amount of labeled data coverage, failing in unfamiliar spatial states. Without relying on extra data collection, our approach learns a recovery policy constructed by an inverse policy inferred from the object keypoint manifold gradient in the original training data. The recovery policy serves as a simple add-on to any base visuomotor BC policy, agnostic to a specific method, guiding the system back towards the training distribution to ensure task success even in OOD situations. We demonstrate the effectiveness of our object-centric framework in both simulation and real robot experiments, achieving an improvement of 77.7\% over the base policy in OOD. Furthermore, we show OCR's capacity to autonomously collect demonstrations for continual learning. Overall, we believe this framework represents a step toward improving the robustness of visuomotor policies in real-world settings. Project Website: \url{https://sites.google.com/view/ocr-penn}\\
\end{abstract}


\section{Introduction}
\label{sec:intro}
Robot learning has achieved significant success in deploying Imitation Learning (IL) methods on real-world robotic systems~\cite{osa2018algorithmic}. One widely studied approach within IL is Behavior Cloning (BC), which has been explored extensively in recent work~\cite{chi2023diffusion, pari2021surprising, zhao2023learning, fu2024mobile, zhu2023viola, shi2023waypoint}. BC methods enable learning control policies directly from demonstrations without the need for explicit environmental modeling, making the process relatively straightforward. However, despite producing promising results, BC is well-known for its susceptibility to the covariate shift problem~\cite{osa2018algorithmic}. This issue arises because traditional BC approaches depend heavily on large quantities of labeled data, which are often obtained through labor-intensive methods such as teleoperation or kinesthetic teaching. Consequently, BC may struggle to perform reliably in out-of-distribution (OOD) scenarios, where data is sparse or noisy—reflecting a broader challenge faced in supervised learning. Addressing this issue typically requires either returning to laborious data collection or utilizing corrective mechanisms, such as guidance from human operators or reinforcement learning (RL) agents~\cite{ross2011reduction, haldar2023teach, kelly2019hg, spencer2020learning}, both of which impose additional deployment efforts on robotic systems.

To enjoy the benefits of strong performing BC policies in distribution (ID) settings while not requiring the human effort of collecting more data or the compute effort of running an RL step when OOD, in this work, we propose a recovery policy framework that brings the system back to the training distribution to ensure task success even when OOD. In particular, we focus on the key challenges of visuomotor policy learning by integrating a recovery policy constructed from the gradient of the training data manifold with a base visuomotor BC policy (e.g., a diffusion policy \cite{chi2023diffusion}). Inspired by the ``Back to the Manifold'' approach~\cite{reichlin2022back}  the recovery policy guides the system back towards the training manifold, at which point the base policy resumes control. However, unlike~\cite{reichlin2022back}, which focuses on recovering from OOD scenarios related to the robot's state, our approach takes an object-centric perspective, specifically addressing OOD situations for task-relevant object states. We believe this object-centric approach significantly enhances the OOD recovery capabilities of visuomotor policies, leading to more robust learning for object manipulation tasks. Furthermore, our recovery framework is designed to be agnostic to the choice of base policy, allowing it to be seamlessly integrated with various BC implementations. This flexibility makes our method adaptable for future developments in imitation learning (IL). In this paper, we make the assumption that we have access to relevant object models. Also, we focus on OOD cases in which the relevant object enters unfamiliar spatial regions.

\textbf{Paper organization} Section \ref{sec:related} presents an overview of the existing works. Section \ref{sec:formulation} describes the problem formulation. Section \ref{sec:method} presents the object-centric recovery policy framework in detail, including its construction of the training data manifold and the keypoint inverse policy. In section \ref{sec:result}, we demonstrate the effectiveness of our approach on several benchmarks, including both simulation and real robot experiments, showing that our recovery policy improves performance when entering unfamiliar states. We also show that our method has the desired property for lifelong learning of visuomotor policies, improving the performance of OOD while not diminishing the in-distribution performance. Section \ref{sec:conclusion} discusses the limitations and future directions.

\section{Related Work}
\label{sec:related}
When deployed to the real world, vision-based IL could easily be initialized or moved to OOD situations, possibly due to bias in data collection and compounding errors. Deploying BC methods OOD could lead to unknown behavior in the low-data region. To address this, a well-known family of approaches is Data Aggregation, which gathers extra data from expert policies (usually provided by humans) through online interaction~\cite{ross2010efficient, ross2011reduction, kelly2019hg, spencer2020learning}. However, performing such an online data collection procedure is an additional burden to the human when building a system. Our method tries to avoid additional online interaction and cumbersome data collection by squeezing as much information from the existing training data. The OOD problem has received much attention from the offline RL community. Offline RL suffers less compounding errors than BC methods as it optimizes for long-term outcomes \cite{levine2020offline}. Yet, still struggles with distribution shifts like extrapolation errors due to limited data. 

To tackle the OOD problem, methods like~\cite{kumar2020conservative, yu2021combo, kostrikov2021offline} try to penalize actions that are far from the data. Moreover, several works~\cite{jiang2024recovering, zhang2022state} also propose to recover back to the training data region, which indirectly shares a similar idea as our work. The paradigm of BC+RL has also been a popular choice for addressing the OOD problem \cite{fujimoto2021minimalist, haldar2023teach, ankile2024imitation}. Our approach takes on a similar direction for training a recovery policy, but instead, we use object-centric BC as the add-on for the base BC policy. Closely related to our work is~\cite{reichlin2022back}, which introduces a vision-based OOD recovery policy by 1) learning an equivariant map that encodes visual observations into a latent space whose gradient corresponds to robot end-effector positions, and 2) following a gradient learned by a Mixture Density Network (MDN)~\cite{bishop1994mixture}. Rather than recovering the robot action, our work focuses on recovering task-relevant objects and inducing the robot action. When dealing with objects, it could be more general to utilize object-centric representations. Many works~\cite{zhu2023viola, chen2024objectcentric, gao2023k, mandikal2021learning, devin2018deep} demonstrate the success of object-centric representation policy learning. In our work, we use keypoints deriving from pose estimation~\cite{wen2024foundationpose} as the object representation for the recovery policy following existing works~\cite{wen2023any, xu2024flow}. 
Recent works also explore using Large Language Models to determine failure and perform recovery~\cite{dai2024racer, cornelio2024recover}.

\begin{figure*}[t] %
    \vspace{4pt}
    \centering
    \includegraphics[trim={0.85cm 9cm 4.5cm 0.85cm},clip,width=0.96\textwidth]{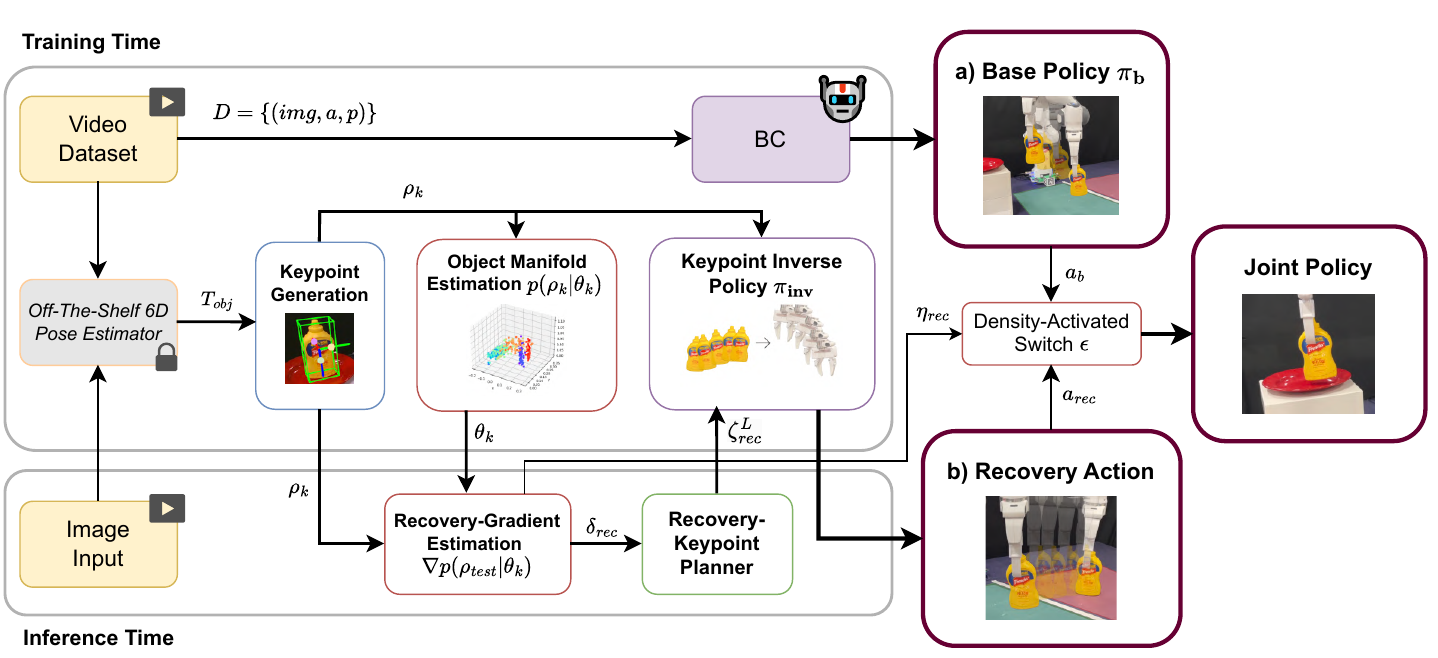} 
    \captionsetup{font=small}
    \vspace{3pt}
    \caption{\textbf{Object Centric Recovery (OCR) Framework.} 
    The OCR Framework augments a base policy $\pi_b$, trained via BC, by returning task-relevant objects to their training manifold, where the base policy takes over. First, we model the distribution of object keypoints in the training data using a Gaussian Mixture Model (GMM). At test time, we compute the gradient of the GMM to derive object-recovery vectors, which are used to plan a recovery trajectory. This trajectory is then converted into robot actions through a Keypoint Inverse Policy $\pi_{inv}$, trained \textit{solely} on the base dataset. Finally, the base policy and the recovery policy are combined into a \textit{joint policy}, allowing seamless interaction between recovery and task execution.}
    \label{fig:ocr-framework} 
    \vspace{-8pt}
\end{figure*}

\section{Recovery Problem Formulation}
\label{sec:formulation}
Distribution shift for learning models is commonly quantified by measuring the Kullback-Leibler (KL) divergence between the distribution of observations during training and the distribution of observations encountered at test time ~\cite{li2023on,hendrycks2020augmix,federici2021an}. This divergence reflects how much the test-time observations deviate from those seen during training, providing a metric to measure if a scenario we encountered is out-of-distribution. Formally, if the probability distribution of the training and testing observations is $P(O)$ and $Q(O)$ respectively, with $O$ representing the set of observations, we say that the testing observation will be considered out-of-distribution (OOD) if,
\begin{equation}
 \mathcal{D}_{KL}(P(O) \Vert Q(O))>\epsilon   
\end{equation} is asserted to be true with some threshold $\epsilon>0$.

We formulate our visuomotor policy interaction with the environment as a Partially Observable Markov Decision Model (POMDP)~\cite{KAELBLING199899}. We describe this POMDP by the tuple $(S,A,O,T,E)$, where $s\in S$ is the set of environmental states, which are directly observable, $a\in A$ is the set of robot actions, and $o\in O$ is the set of visual observations. The transition function $T:S\times A\rightarrow S$ dictates how the unobservable state changes when robot actions are performed, and the emission function $E:S\rightarrow O$ is a surjective function that determines the visual observations given states. 

Given this formulation, we can reformulate the KL Divergence OOD metric as follows, 
\begin{equation}
\mathcal{D}_{KL}(P(E(S)) \Vert Q(E(S)))>\epsilon . 
\end{equation}
Hence, fundamentally, given an observation-level out-of-distribution scenario, if the environmental state variables are recovered back into the training distribution, the observations will also be recovered back into distribution. However, the recovery of \textit{all} state variables is difficult to tackle all at once under the imitation learning framework, which typically has access to only task-relevant demonstrations. Therefore, for this work, we specifically focus on the recovery of task-relevant objects in manipulation tasks. Unlike previous data aggregation or reinforcement learning approaches, we aim for our recovery framework to exclusively leverage the training demonstrations of the base policy and not require any additional policy-related data collection.

\section{Method}
\label{sec:method}

We present our approach in augmenting a \textit{base} policy trained via Behavior Cloning (BC) by incorporating an \textit{object-centric} recovery strategy, which enables task-relevant objects to return to its Euclidean training manifold where the base BC policy functions at its best. For our work, we will assume task-relevant objects in the scene are rigid and non-deformable.

To achieve this, we introduce the Object-Centric Recovery (OCR) framework, as illustrated in Figure \ref{fig:ocr-framework}. We first explicitly model the distribution of objects keypoints in the training dataset with a Gaussian Mixture Model (GMM)~\cite{bishop2006pattern} $p(\rho_{k,t}^{(i)} | \theta_k) = \sum^M_{m=1}\lambda_{k,m} \mathcal{N}_{\theta_k}(\rho_{k,t}^{(i)}|\mu_{k,m},\Sigma_{k,m})$, where $\rho_{k,t}^{(i)}$ are keypoints in the dataset and $\theta_k = \{(\lambda_{k,m}, \mu_{k,m},\Sigma_{k,m})\}^M_{m=1}$ parameters of the GMM (Section~\ref{method_keypoint}, \ref{method_manifold}). At test time, we evaluate the gradient $\nabla p(\rho^{test}_k|\theta_k)$ 
to obtain the object-recovery vectors, which we use to plan for an object-recovery trajectory $\zeta_{rec}^L$ (Section \ref{method_object_vec} and \ref{method_object_plan}). We then translate this trajectory into robot actions via a Keypoint Inverse Policy $\pi_{inv}$ that is trained using the base dataset (Section \ref{method_inv}). Lastly, Section \ref{method_joint_policy} describes how the base policy interacts with the recovery policy to become the OCR \textit{joint policy}.

\subsection{Base BC Policy}\label{method_base}
Our formulation considers a generic visuomotor policy that outputs future actions based on past visual observations as the base BC policy. We consider such a liberal formulation to demonstrate that our framework can work alongside any variations of BC policy. Formally, we define a typical visuomotor policy training dataset as $\mathbf{D}_b=\{\mathbf{d}^{(i)}_b\}^N_{i=1}$, where each episode $\mathbf{d}^{(i)}_b=\{(\mathbf{o}^{(i)}_t,\mathbf{a}^{(i)}_t, \mathbf{p}^{(i)}_t)\}^T_{t=1}$ consists of the observations $\mathbf{o}^{(i)}_t$, robot actions $\mathbf{a}^{(i)}_t$, and robot proprioception $\mathbf{p}^{(i)}_t$ at time step $t$. Then, under the imitation learning framework, a base visuomotor policy $\pi_b$ that is parameterized by $\phi_b$ is learned by optimizing the following behavior cloning objective: 
\begin{equation}
\pi_b^*=\arg\min_{\theta_b}E_{(\mathbf{o},\mathbf{a}, \mathbf{p})\sim\mathbf{D}_b}\left[\mathcal{L}\left(\pi_b(\mathbf{o},\mathbf{p}),\mathbf{a}\right)\right]
\end{equation} 
Where the loss function $\mathcal{L}$ is typically Cross-Entropy Loss or Mean-Squared Error.

\subsection{Object-Centric Recovery Policy}
\subsubsection{\textbf{Keypoint Generation}}\label{method_keypoint}
We choose to use artificial object keypoints to represent object poses for studying object-centric recovery, as keypoints allow us to tightly couple the position and orientation of the object, facilitating a more accurate estimation of its distribution during training.

We consider the same visuomotor policy training dataset formulation $\mathbf{D}_b=\{\mathbf{d}^{(i)}_b\}^N_{i=1}$, where each episode $\mathbf{d}^{(i)}_b=\{(\mathbf{o}^{(i)}_t,\mathbf{a}^{(i)}_t, \mathbf{p}^{(i)}_t)\}^T_{t=1}$ consists of the observations $\mathbf{o}^{(i)}_t$, robot actions $\mathbf{a}^{(i)}_t$, and robot proprioceptions $\mathbf{p}^{(i)}_t$ at time step $t$. To extract object poses from these visuomotor datasets, we employ off-the-shelf object pose estimators (e.g. ~\cite{Hai2023Shape,wen2024foundationpose}) to transform each observation frame $\mathbf{o}^{(i)}_t$ into the object pose $\mathbf{T}^{(i)}_{obj,t}$. Next, we define an arbitrary set of keypoints $\mathbf{P}=\{p_k\}^n_{k=1}$, where each keypoint $p_k\in\mathbb{R}^d$. For each keypoint $p_k$ at time step $t$ in demonstration $i$, we compute the transformed keypoints $\rho_{k,t}^{(i)} = h^{-1}(\mathbf{T}^{(i)}_{obj,t} h(p_k))$, where $h$ represents the function that converts points into homogeneous coordinates. The transformed keypoint $\varrho_t^{(i)}=\{\rho_{k,t}^{(i)}\}^n_{k=1}$ then serves as the keypoint representation of the object’s current pose. Thus, using $\mathbf{D}_b$, we create a new dataset that will be used for recovery $\mathbf{D}_{rec}=\{\mathbf{d}^{(i)}_{rec}\}^N_{i=1}$, where each episode $\mathbf{d}^{(i)}_{rec}=\{(\varrho_t^{(i)},\mathbf{T}^{(i)}_{obj,t},\mathbf{a}^{(i)}_t,\mathbf{p}^{(i)}_t)\}^T_{t=1}$ consists of the keypoints, object poses, robot actions and proprioception at each time step. 
\vspace{3pt}
\subsubsection{\textbf{Object Manifold Estimation}}\label{method_manifold}
To estimate the manifold of the object distribution in the training dataset, we fit a Gaussian Mixture Model (GMM)~\cite{bishop2006pattern} on each keypoint using its positions across every time step in every demonstration. Specifically, given dataset $\mathbf{D}_{kp,k}=\left\{\{(\rho_{k,t}^{(i)})\}^T_{t=1}\right\}^N_{i=1}$ consisting of one object keypoint $k$ across all time step $t$ in every demonstration $i$, we model the probability of each $\rho_{k,t}^{(i)}$ as a weighted sum of $M$ Gaussian distributions: 
\begin{equation}
p(\rho_{k,t}^{(i)} | \theta_k) = \sum^M_{m=1}\lambda_{k,m} \mathcal{N}_{\theta_k}(\rho_{k,t}^{(i)}|\mu_{k,m},\Sigma_{k,m}),
\end{equation}

where $\lambda_{k,m}$ is the mixing coefficient of keypoint $k$ for the $m$-th Gaussian, $\mathcal{N}(\rho_{k,t}^{(i)}|\mu_{k,m},\Sigma_{k,m})$ is the Gaussian probability density function of keypoint $k$ for the $m$-th component with mean $\mu_{k,m}$ and covariance $\Sigma_{k,m}$, and $\theta_k = \{(\lambda_{k,m}, \mu_{k,m},\Sigma_{k,m})\}^M_{m=1}$ are the parameters of the model that estimates the distribution of keypoint $k$. To fit this GMM, we used Expectation-Maximization~\cite{bishop2006pattern} to maximize the likelihood estimation of the model on the data. Computing a GMM for all $n$ keypoints would result in parameters $\Theta = \{\theta_k\}^n_{k=1}$ that collectively estimate the probability distribution of the object keypoints.
\vspace{3pt}
\subsubsection{\textbf{Keypoint Inverse Policy}}\label{method_inv}
To facilitate object manipulation for recovery, we propose the use of a Keypoint Inverse Policy $\pi_{inv}$, which is designed to translate a sequence of object-keypoint trajectory along with the robot's current state into the corresponding robot actions necessary to execute those object motions effectively. Formally, if we define $K$ to be the set of object keypoints observed and $P \subseteq S$ to be the set of robot proprioception states, then $\pi_{inv}: K^L \times P \rightarrow A^L$, where $L$ is the observation length. We utilize the dataset of object keypoints, pose, and action tuples $\mathbf{D}_{rec}=\{\{(\varrho_t^{(i)},\mathbf{T}^{(i)}_{obj,t},\mathbf{a}^{(i)}_t,\mathbf{p}^{(i)}_t)\}^T_{t=1}\}^N_{i=1}$ that we described in Section \ref{method_keypoint} to directly train $\pi_{inv}$ with a imitation learning objective. We do this by pulling sequences of length $L$ from the keypoint and action datasets to form $\{\varrho_t^{(i)}\}^{j+L}_{t=j}$ and $\{\mathbf{a}_t^{(i)}\}^{j+L}_{t=j}$, and the initial proprioception of the sequence $\mathbf{p}^{(i)}_j$. For simplicity, we will name these quantities $\varrho_{org}^L$, $\mathbf{a}_{org}^L$, $\mathbf{p}_{org}$ respectively. Thus, we end up with the following training objective: 
\begin{equation}
\pi_{inv}^*=\arg\min_{\theta_{inv}}E_{\left(\varrho_{org}^L,\mathbf{a}_{org}^L,\mathbf{p}_{org}\right)\sim\mathbf{D}_{rec}}\left[\mathcal{L}\left(\pi_{inv}\left(\varrho_{org}^L, \mathbf{p}_{org}\right),\mathbf{a}_{org}^L\right)\right]
\end{equation} 
However, by training this objective directly, we will still run into the same issue of distribution shift, having no keypoints-to-action coverage on the OOD regions to generate properly useful manipulation outputs. To alleviate this, we propose the use of the initial object pose $\mathbf{T}^{(i)}_{obj,t}$ to "zero-out" the data sequence. Specifically, instead of using the original sequence, we use the initial object pose of each sequence $\mathbf{T}^{(i)}_{obj,j}$ to modify the sequence into $\{(\mathbf{T}^{(i)}_{obj,j})^{-1}\varrho_t^{(i)}\}^{j+L}_{t=j}$, $\{(\mathbf{T}^{(i)}_{obj,j})^{-1}\mathbf{a}_t^{(i)}\}^{j+L}_{t=j}$, and $(\mathbf{T}^{(i)}_{obj,j})^{-1}\mathbf{p}^{(i)}_j$. For simplicity, we will name these quantities $\varrho_{\sim0}^L$, $\mathbf{a}_{\sim0}^L$, $\mathbf{p}_{\sim0}$ respectively. Hence, the learning objective becomes: 
\begin{equation}\pi_{inv}^*=\arg\min_{\theta_{inv}}E_{(\varrho_{\sim0}^L,\mathbf{a}_{\sim0}^L,\mathbf{p}_{\sim0})\sim\mathbf{D}_{rec}}\left[\mathcal{L}\left(\pi_{inv}\left(\varrho_{\sim0}^L, \mathbf{p}_{\sim0}\right),\mathbf{a}_{\sim0}^L\right)\right]
\end{equation}
In other words, the keypoint inverse policy only needs to learn to output robot actions from object keypoint trajectories that initialize from the identity frame. At test time, to carry out an object motion, we input a desired keypoint trajectory in the current object frame, and the policy outputs the corresponding robot action in that same frame. To execute the robot action, we then transform the action from the object frame to the robot frame. 
This way, regardless of the object and robot end-effector's true pose in Euclidean space, OOD or ID, as long as we have access to the current object pose, we can output robot actions that are useful for manipulation. 
In addition, we can think of this as a way of ``compressing" the input domain of the keypoint inverse policy based on the information available to us (object pose), making the learning problem extremely data efficient.
\vspace{3pt}
\subsubsection{\textbf{Object-Recovery Vectors}}\label{method_object_vec}
At test time, we obtain the explicit current pose of the recovery object via pose estimation and generate keypoints using the same methodology as described in Section \ref{method_keypoint}, obtaining $\mathbf{\varrho}^{test}=\{\rho^{test}_{k}\}^n_{k=1}$. For each keypoint, we build a computation graph of the probability density function of the GMM with parameters $\theta_k$ to output the probability density $\eta^{test}_k$ with respect to $\rho^{test}_{k}$. With this, we used automatic differentiation to output the gradient vector $\delta^{test}_k = \nabla p(\rho^{test}_k|\theta_k)$. However, the norm of this gradient vector $\Vert\delta^{test}_k\Vert$ is strictly non-negative and increases as $\rho^{test}_{k}$ approaches regions of increasingly higher density parameterized by the GMM with parameters $\theta_k$, which is in contrast with how we want recovery to take place - to approach recovery faster when the object is further away, and slower when the object is closer. To solve this, we modify the magnitude of $\delta^{test}_k$ by a monotonically decreasing function, the \textit{parameterized negative exponential} function, which we define as $q(x)=e^{\frac{\phi-x}{\eta}}$. Thus, the modified recovery gradient is 
\begin{equation}
\delta^{mod}_k=q(\Vert\delta^{test}_k\Vert)\frac{\delta^{test}_k}{\Vert\delta^{test}_k\Vert} 
\end{equation}
Since the recovery gradient and density would differ for each keypoint $k$, we will use the mean gradient and mean density for the final recovery policy, given by:
\begin{equation}
\delta_{rec} = \sum^{n}_{k=1}\frac{\delta^{mod}_k}{n}, \ \eta_{rec} = \sum^{n}_{k=1}\frac{\eta^{test}_k}{n}
\end{equation}
Hence, at each time step during test time, we output an object recovery tuple $(\delta_{rec},\eta_{rec})$.
\vspace{3pt}
\subsubsection{\textbf{Recovery Keypoint Planner.}} \label{method_object_plan} From the object recovery vector $\delta_{rec}$, we can generate a naive recovery keypoint trajectory like so: 
\begin{equation}
\left[\{t\alpha\delta_{rec}+\rho^{test}_{k}\}^n_{k=1}\right]^L_{t=1}
\end{equation}
where $\alpha$ is a scaling hyperparameter that we can tune at test time to optimize for the trajectory step size.
However, this formulation does not take into account the feasibility of executing such a trajectory, which is paramount in ensuring the quality of the recovery. To this end, we propose a heuristic planner, using the distance of the position between the robot end-effector and the object pose as a heuristic for how much ``delay" is added to the object trajectory before it starts moving, thus providing the robot with enough time to approach the object for manipulation. Specifically, we will define a maximum and a minimum distance where the object can be effectively manipulated, which we denote as $d_{\max}$ and $d_{\min}$, which can be tuned easily at test time. Then, we simply fit a linear function between points $(d_{\min}, L)$ and $(d_{\max}, 0)$, and clip the range between $[L,0]$. Formally, if we denote the norm between the position of the end-effector and object as $d_{pos}$, then the delay function is written as:
\begin{equation}
 df(d_{pos})=\min(\max(0,\lfloor\frac{-L}{d_{max}-d_{min}} * (d_{pos}-d_{min}) + L\rceil)L)   
\end{equation}
Our proposed keypoint recovery trajectory is expressed as:
\begin{equation}
\zeta_{rec}^L = \left[\{\max(0,t-df(d_{pos}))\alpha\delta_{rec}+\rho^{test}_{k}\}^n_{k=1}\right]^L_{t=1}
\end{equation}
\subsubsection{\textbf{Final Policy}}\label{method_joint_policy}
We join our base policy and recovery action as a \textit{Joint Policy} via a density-activated switch. Specifically, after computing the mean keypoint density $\eta_{rec}$, we use a tunable hyperparameter $\epsilon_{rec}$ to define the threshold for distinguishing between OOD and ID scenarios. If the scenario is classified as OOD, the recovery pipeline is activated; otherwise, the base policy will proceed with the standard BC process. Formally, our joint policy is: 
\begin{equation} 
\mathbf{\pi_{joint}} = \mathbb{I}_{\{ \eta_{rec} \geq \epsilon_{rec} \}}\pi_b(\mathbf{o_t},p_t) + \mathbb{I}_{\{ \eta_{rec} < \epsilon_{rec}
\}}\pi_{inv}(\zeta_{rec}^L, p_t)
\end{equation}
Algorithmically, we can summarize our Joint Policy in Algorithm \ref{algo_joint-policy}:

\begin{algorithm} \label{app:joint_policy_algo}
\caption{Joint Policy Algorithm}
\begin{algorithmic}[1] \label{algo_joint-policy}
\STATE Initialize $\pi_b$, $\pi_{inv}$, GMMs with parameters $\Theta$
\WHILE{Task not done}
    \STATE Collect observation $\mathbf{o}_t$, proprioception $p_t$. Compute object pose $\mathbf{T}_{obj,t}$, keypoints $\varrho_t$.
    \STATE Evaluate mean keypoint recovery vector $\delta_{rec}$ and mean keypoint density $\eta_{rec}$.
    \IF{$\eta_{rec} < \epsilon_{rec}$} 
        \STATE Compute keypoint recovery trajectory $\zeta_{rec}^L$
        \STATE Compute recovery action trajectory $\mathbf{a}_{out}^L = \pi_{inv}(\zeta_{rec}^L, p_t)$
    \ELSE
        \STATE Compute base action trajectory $\mathbf{a}_{out}^L = \pi_b(\mathbf{o}_t, p_t)$
    \ENDIF
    \FOR{$a$ in $\mathbf{a}_{out}^L$}
        \STATE Execute action $a$
    \ENDFOR
\ENDWHILE
\end{algorithmic}
\end{algorithm}

\section{Evaluation}
\label{sec:result}
We systematically evaluated the Object-Centric Recovery framework's capabilities on (i) a simulated 2D non-prehensile task, (ii) a simulated 3D prehensile task, and (iii) a real-robot prehensile task. We selected these scenarios to demonstrate the framework's versatility and robustness in handling a wide range of manipulation settings. Each task scenario has a designated in-distribution (ID) region where demonstration data exists, and an out-of-distribution (OOD) region void of demonstration data. We carefully evaluated our recovery policy's effectiveness in both the ID and OOD regions against a baseline policy. 

To the best of our knowledge, given the absence of existing methods for object-centric recovery in visuomotor policies, we benchmarked our results against the OOD performance of the base BC policy. To ensure consistency across all tasks, we employed the vision-input U-Net-based diffusion policy~\cite{chi2023diffusion} as our base BC policy, and the low-dimensional diffusion policy was utilized as the architecture for our keypoint inverse policy. This base BC policy was used both as the baseline for evaluation as well as the base policy for our OCR joint policy. Our results show that, compared to the baseline, the Object-Centric Recovery framework consistently achieved a high task-completion rate in object OOD scenarios, with an average success rate of \textbf{81.0\% across the three evaluated tasks}, which is an improvement of \textbf{77.7\%} over the base policy in OOD.

In addition, we show that in a life-long continual learning scenario, we can employ the OCR framework to \textit{automate demonstration collection} for OOD scenarios. We show that when the OCR-collected demonstrations are augmented alongside the original base policy's training dataset for incremental learning, it can imbue the improved base policy with the ability to recover at a high success rate without diminishing its performance in the original ID regions.

\subsection{Experimental Setups}
\begin{figure}[t]
    \vspace{8pt}
    \centering
    \includegraphics[width=\linewidth]{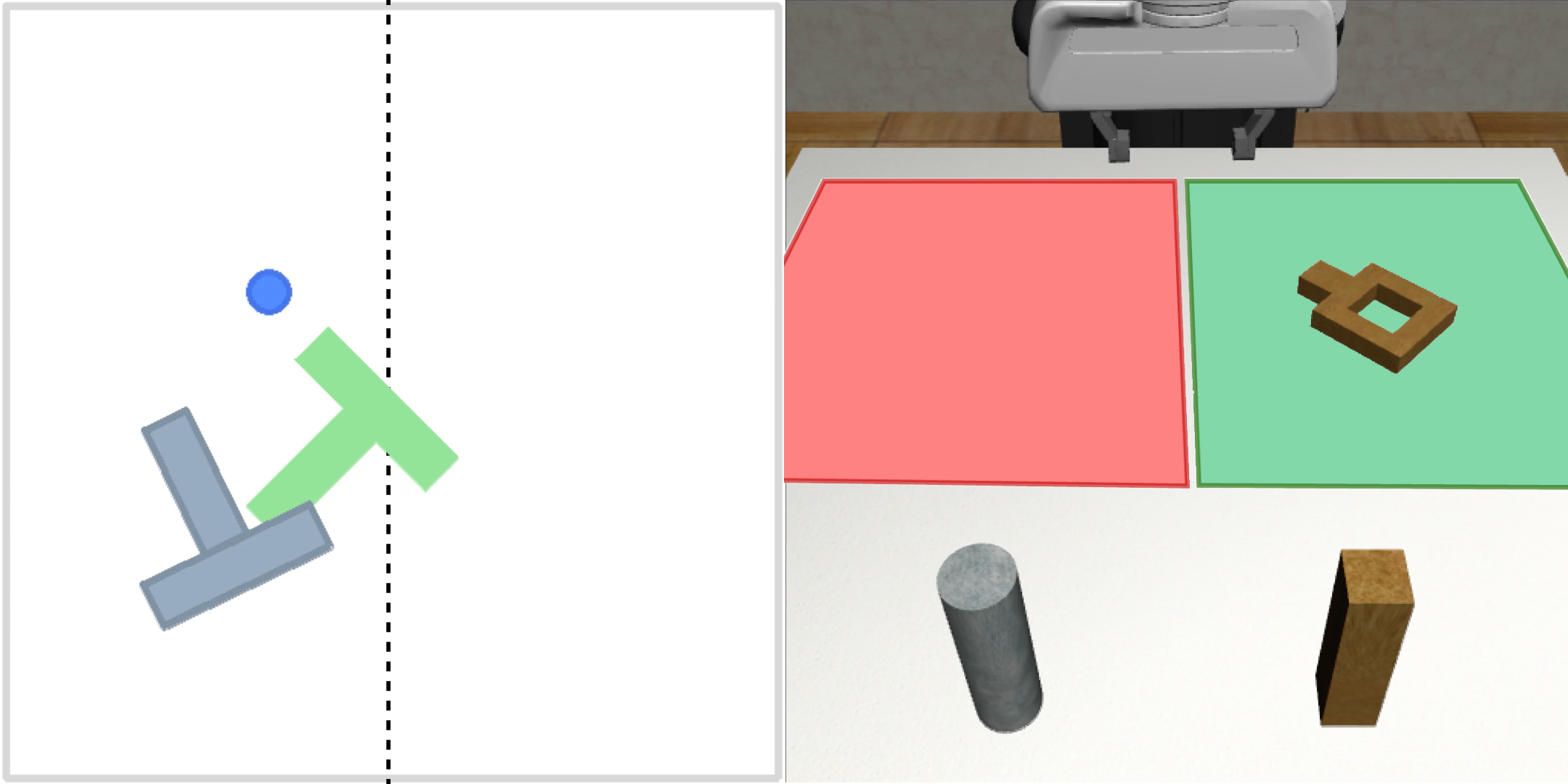}
    \captionsetup{font=small}
    \caption{(\textbf{Left}) shows the Push-T Task's ID and OOD regions divided by a dashed line. (\textbf{Right}) shows the Square Task's ID and OOD regions drawn out by the green and red regions, respectively.}
    \label{fig:pusht-square}
\end{figure}

\textit{Experiment 1)} \textbf{Non-Prehensile, Sim, Push-T Task~\cite{Florence2021ImplicitBC}}: This 2D simulated task involves pushing a T-shaped block (gray) toward a fixed target using a circular end-effector agent (blue). At each reset, both the initial pose of the T block and the initial position of the end-effector are randomized. The task is particularly challenging due to the requirement for discontinuous, non-linear end-effector actions. To highlight the OCR framework’s capability to handle such complexity, we provided demonstrations initialized exclusively on the \textit{left} side of the screen, as shown by the dashed line separating the screen in Figure \ref{fig:pusht-square}. This setup designates the left side as ID and the right as OOD. For PushT, we recorded 100 demonstrations in the ID region. 

\vspace{1mm}
\textit{Experiment 2)} \textbf{Prehensile, Sim, Robomimic Square Task~\cite{robomimic2021}}: This simulated task requires the robot to pick up a notched square-shaped object with a hole in the middle, transport it, and drop it through a fixed square peg. The initial pose of the square object is randomized within the $\mathrm{SE}(2)$ space on the table, and the initial position of the end-effector is also randomized. We used Robomimic's PH dataset, which is 200 demonstrations initialized exclusively on the \textit{right} side of the table (ID), as shown by the green-shaded region in Figure \ref{fig:pusht-square}. Hence, the left side of the table, or the red-shaded region, is considered OOD.

\begin{figure}[t]
    \vspace{8pt}
    \centering
    \includegraphics[trim={5.0cm 0cm 5.0cm 0cm},clip,width=\linewidth]{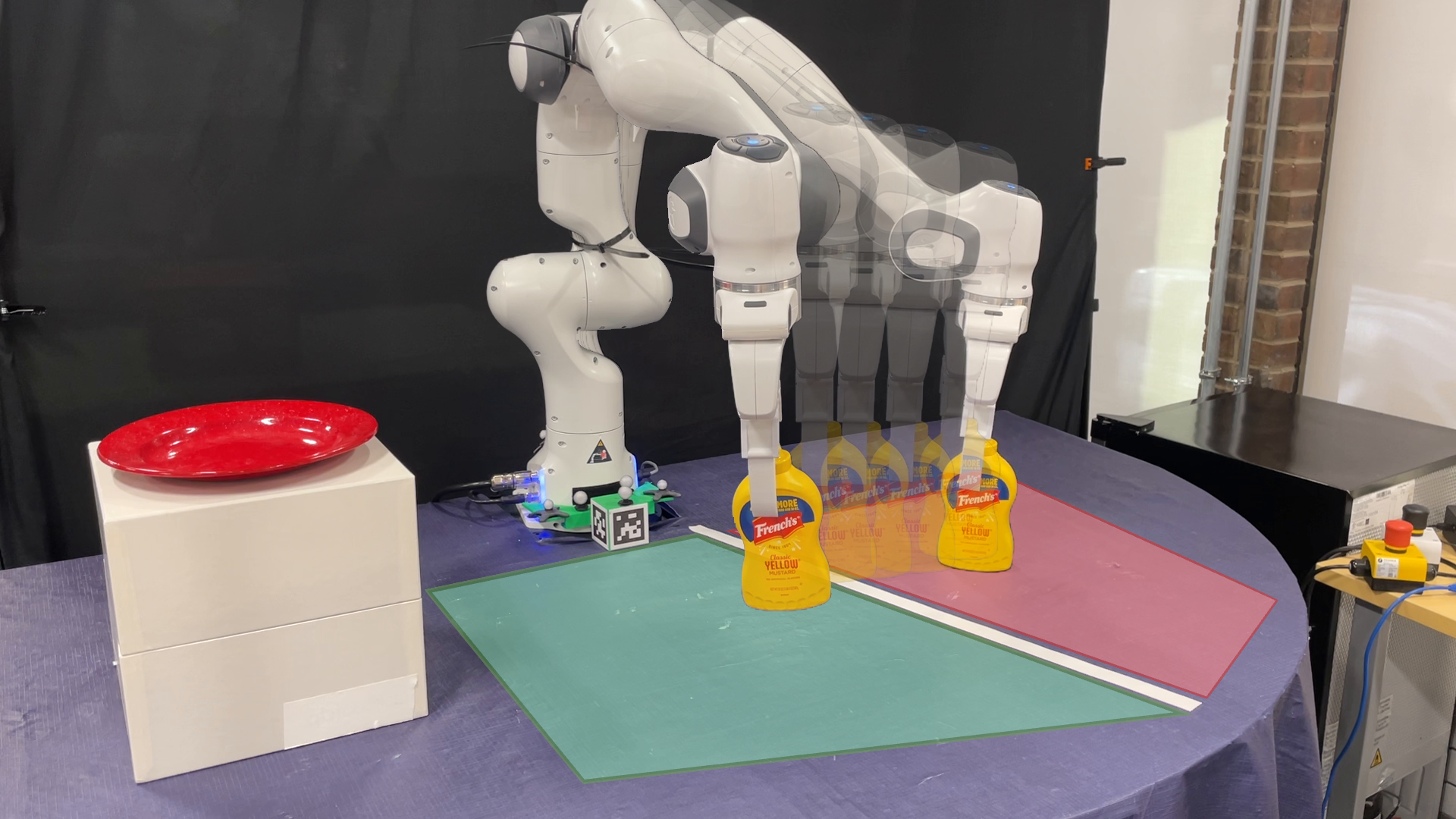}
    \caption{(\textbf{Right}) shows our Franka recovering from OOD (red-shaded) to ID (green-shaded) in the Bottle Task.}
    \label{fig:bottle}
\end{figure}

\vspace{1mm}
\textit{Experiment 3)} \textbf{Prehensile, Real, Bottle Task}: The objective of the bottle task is for the robot to grasp a yellow bottle, transport it, and place it onto an elevated red plate. The initial pose of the bottle is randomized within the $\mathrm{SE}(2)$ space on the table, and the initial position of the end-effector is also randomized. As illustrated in Figure \ref{fig:bottle}, the green-shaded region represents the demonstrated ID region, while the red-shaded area indicates the OOD region. We provided 115 demonstrations in the green-shaded region for policy training. We used a Franka Panda robot and Polymetis~\cite{Polymetis2021} as the controller interface. We used Foundation Pose~\cite{wen2024foundationpose} for 6D object pose estimation and Grounded-SAM~\cite{ren2024grounded} to obtain the object mask.

\begin{table}[t]
    \centering
    \small
    \begin{tabular}{l|cc|c}
    \toprule
     & \multicolumn{2}{c|}{Base Policy} & Joint Policy (Ours) \\
     & ID & OOD & OOD \\
    \midrule
    Push-T & 0.90 & 0.10 & \textbf{0.93} \\
    Square & 0.87 & 0.00 & \textbf{0.80} \\
    \bottomrule
    \end{tabular}
    \caption{Simulated task success rate of the base policy vs. joint policy in OOD scenarios, with ID scenario baseline as the base policy.}
    \label{tab:sim}
\end{table}

\begin{table}[t]
    \centering
    \small
    \begin{tabular}{l|cc|c}
    \toprule
     & \multicolumn{2}{c|}{Base Policy} & Joint Policy (Ours) \\
     & ID & OOD & OOD \\
    \midrule
    Bottle & 0.60 & 0.00 & \textbf{0.70} \\
    \bottomrule
    \end{tabular}
    \caption{Real task success rate of the base policy vs. joint policy in OOD scenarios, with ID scenario baseline the base policy.}
    \label{tab:real}
\end{table}

\begin{table}[t]
    \vspace{8pt}
    \centering
    \small
    \begin{tabular}{l|cc|cc}
    \toprule
     & \multicolumn{2}{c|}{Org Base Policy} & \multicolumn{2}{c}{Aug Base Policy} \\
     & ID & OOD & ID & OOD \\
    \midrule
    Push-T & 0.90 & 0.10 & 0.97 & \textbf{0.80} \\
    \midrule
    Square & 0.87 & 0.00 & 0.87 & \textbf{0.76} \\
    \bottomrule
    \end{tabular}
    \caption{Simulated task success rate of the original base policy vs. augmented base policy (trained with additional OCR generated data) in ID and OOD scenarios.}
    \label{tab:continue}
\end{table}

\vspace{1mm}
\textit{Experiment 4)} \textbf{Continual Learning in Simulation}: To demonstrate the OCR framework’s effectiveness in lifelong continual learning, we used OCR to autonomously collect data in OOD regions for incremental training. We initalize the environment OOD and let the policy roll out actions until some predefined $\epsilon_{rec}$ threshold for reaching ID was met. During the rollout, we recorded the standard observations and proprioception as augmented demonstrations. From 100 OOD initializations, we collected this augmented demo dataset $\mathbf{D_{aug}}$, which we appended to $\mathbf{D_b}$ directly to resume training on the base policy checkpoint. We did this for both the Push-T and Robomimic Square tasks and evaluated the resulting \textit{augmented base policy} in both their original ID and OOD scenarios.

\subsection{Experimental Results \& Analysis}
1) \textbf{Push-T}. As shown in Table \ref{tab:sim}, the Push-T base policy generalized to the OOD region object initialization in only 10\% of cases, mainly due to random accidental actions. In contrast, using OCR, recovery actions are able to intentionally interact with relevant objects in locally demonstrated ways (e.g., end-effector circling the T-shape to position itself correctly for pushing) even in OOD regions. In addition, the recovery actions were able to bring the T-shape back into the ID region very reliably, where the base policy completes the task. Across 30 random OOD initializations, the OCR framework achieved a 93\% task success rate, significantly outperforming the baseline.

\vspace{1.5mm}
2) \textbf{Square}. In object OOD scenarios, the Square base policy consistently attempted to move the end-effector toward the direction of the object but never reached it, resulting in no successful task completions. In contrast, as shown in Table \ref{tab:sim}, the OCR framework was able to execute grasping the object OOD, manipulating the object for recovery, and allowing the base policy to take over ID reliably. We observed an overall 80\% task success rate in OOD scenarios for this task, which is a substantial improvement over the base policy.

\vspace{1.5mm}
3) \textbf{Bottle}. For the real bottle task, we observed the Bottle base policy failed similarly to the Square base policy when the object is OOD. However, as shown in Table \ref{tab:real}, the OCR framework effectively handled the recovery for the bottle task, achieving a 70\% task success rate in OOD scenarios, significantly outperforming the base policy. Interestingly, we observed that the OCR joint policy's OOD success rate is significantly higher than the base policy's ID success rate for the bottle task. We hypothesize that this can be attributed to the OCR framework's ability to move objects toward regions of high training density regardless of where the objects are initialized.

\vspace{1.5mm}
4) \textbf{Continual Learning}. By resuming training on the base policy with the augmented dataset $D_{aug}$ that is autonomously collected via the OCR framework, we enabled the augmented policy to recover independently in both of the previously tested simulated tasks. On Push-T, the augmented policy achieved 80\% task completion in the original OOD regions, as shown in Table \ref{tab:continue}, while on 
 the Square task, it achieved an OOD task completion rate of 76\%. Both augmented policies showed significant improvements over their base counterparts while not sacrificing their performance in the original ID scenarios; in fact, the augmented policy in the Push-T task showed enhanced performance in ID as well, improving from 90\% to 97\%. We hypothesize this is due to the OCR's augmented dataset consistently providing robot actions that move the object toward regions of higher density, even on the ID side. In other words, OCR demonstrations likely offer actions that \textbf{converge} the object to the base policy, complementing it rather than replacing it. We believe that this showcases the OCR framework's ability to provide valuable data for continual learning.

\section{Conclusion \& Future Work}
\label{sec:conclusion}
In this work, we proposed the Object-Centric Recovery policy framework designed to address out-of-distribution challenges in visuomotor policy learning, by recovering task-relevant objects into distribution without requiring additional data collection. When our framework was tested against various manipulation tasks and environments, it demonstrated considerable improvement in performance in OOD regions. Furthermore, the framework's capacity for continual learning highlights its potential to autonomously enhance policy behavior over time. 

However, there are a few key limitations to our approach. First, our reliance on explicit object poses restricts its applicability to articulated and deformable objects. Furthermore, our use of state-based distribution manifold estimation is, at best, only a proxy of the true visual distribution of visuomotor policies. In addition, the 3D keypoint representation may be inconsistent across training and inference, which our method heavily relies on. We would like to address this for future work. Finally, future works can extend OCR by incorporating more flexible scene representations to recover from a broader range of OOD scenarios at higher accuracy. Despite these limitations, we believe that our framework represents a step toward improving the robustness of visuomotor policies in real-world settings. 


\bibliographystyle{IEEEtran}
\normalsize
\bibliography{references}

\end{document}